\documentclass[extendedabs]{recpad2k}
\usepackage[named]{algo}
\usepackage{graphics}
\usepackage{graphicx}
\usepackage{amssymb}
\usepackage{amsmath}
\usepackage{subfigure}
\usepackage{paralist}

\usepackage{url}
\title{A novel sparsity and clustering regularization}

\addauthor{Xiangrong Zeng}{Xiangrong.Zeng@lx.it.pt}{1}
\addauthor{M\'{a}rio A. T. Figueiredo}{mario.figueiredo@lx.it.pt}{1}

\addinstitution{
Instituto de Telecomunica\c{c}\~oes\\
 Instituto Superior T\'{e}cnico\\
 Lisbon, Portugal
}

\runninghead{Student, Prof, Collaborator}{RECPAD Author Guidelines}

\begin{document}

\maketitle

\begin{abstract}
We propose a novel {\it SPARsity and Clustering} (SPARC) regularizer, which
is a modified version of the previous {\it octagonal shrinkage and
clustering algorithm for regression} (OSCAR), where, the proposed regularizer consists of a $K$-sparse constraint and a pair-wise $\ell_{\infty}$ norm restricted on the $K$ largest components in magnitude. 
The proposed regularizer is able to separably enforce $K$-sparsity and encourage the non-zeros to be equal in magnitude. Moreover, it can accurately group the features without shrinking their magnitude. In fact, SPARC is closely related to OSCAR, so that the proximity operator of the former can be efficiently computed based on that of the latter, allowing using proximal splitting algorithms
to solve problems with SPARC regularization. Experiments on synthetic data and with benchmark breast cancer data show that SPARC is a competitive
group-sparsity inducing regularizer for regression and classification.

\end{abstract}

\section{Introduction}
\begin{figure}
	\centering
		\includegraphics[width=0.65\columnwidth]{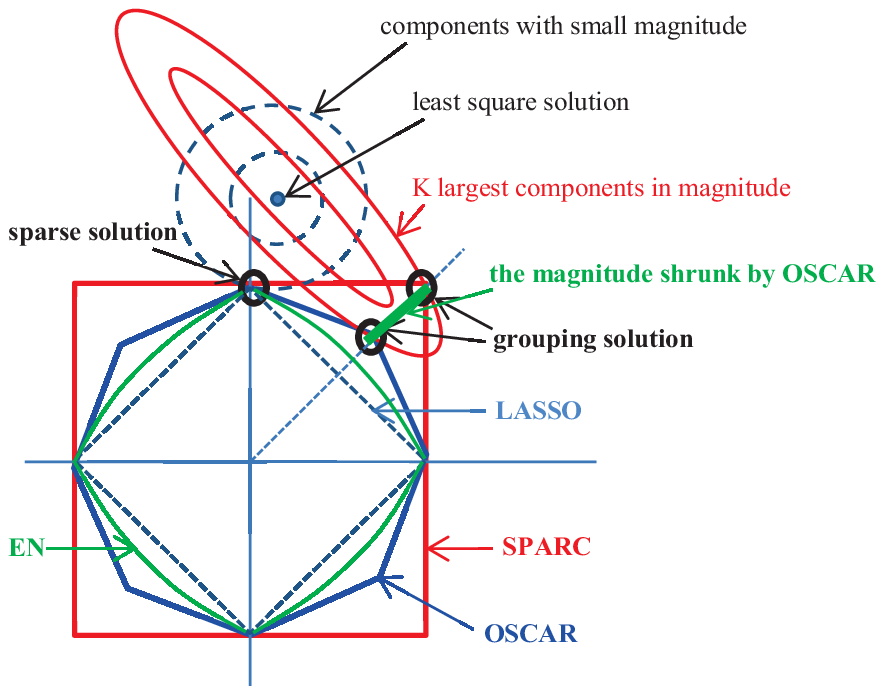}
	\caption{Demonstration of different regularizers}
	\label{fig:regularizers}
\end{figure}

 In recent years, much attention has been paid not only to sparsity but also to structured/group
 sparsity. Several group-sparsity-inducing regularizers have been proposed, including {\it group LASSO}
(gLASSO) \cite{yuan2005model}, {\it fused LASSO} (fLASSO) \cite{tibshirani2004sparsity},
{\it elastic net} (EN) \cite{zou2005regularization}, {\it octagonal shrinkage and
clustering algorithm for regression} (OSCAR) \cite{bondell2007simultaneous}, and several others,  not listed here due to space limitations
(see review in \cite{bach2012structured}).
However, gLASSO (and its many variants and descendants \cite{bach2012structured}) require
prior knowledge about the structure of the groups, which
is a strong requirement in many applications, while fLASSO depends on a given order of variables;
these two classes of approaches are thus better suited to signal processing applications
than to variable selection and grouping in machine learning problems, such as
regression or classification (where the order od the variables is often meaningless).
In contrast,  EN and OSCAR were proposed for regression
problems and do not rely on any ordering of the variables or knowledge about group structure.
The OSCAR regularizer (shown in \cite{zhong2012efficient} to outperform
EN in feature grouping) is defined as
\[
\phi_{\mbox{\tiny OSCAR}}^{\tiny{\lambda_1, \lambda_2}}\left({\bf x}\right) =
\lambda_1 \left\|{\bf x}\right\|_1 + \lambda_2 \sum_{i<j}\max\left\{\left| x_i\right|,
\left|  x_j\right| \right\}
\]
 where $\lambda_1 $ and $\lambda_2$ are non-negative
parameters (which, in practice, can be obtained, for example, 
by {\it cross validation})  \cite{zhong2012efficient}.
The $\ell_1$ norm and the pairwise $\ell_\infty$ penalty simultaneously encourage
the components to be sparse and equal in magnitude, respectively. However, it may happen that
components with small magnitude that should be shrunk to zero by the $\ell_1$
norm are also penalized by the pairwise $\ell_\infty$ term, which may prevent accurate grouping;
moreover, components with large magnitude that should simply be grouped by the pairwise $\ell_\infty$ norm
are also shrunk by the $\ell_1$ norm (see Figure \ref{fig:regularizers}). In this paper, to overcome these drawbacks,
we propose  the {\it SPARsity-and-Clustering} (SPARC) regularizer,
where the cardinality of the support of the solution is restricted and the pairwise
$\ell_{\infty}$ penalty is applied only to the non-zero elements (see Figure \ref{fig:regularizers}).
We also show how to compute the proximity operator of the SPARC regularizer, which allows using proximal splitting algorithms to problems with this regularizer.

\section{Proposed Formulation and Approach}
\label{sec:problem_formulation}
A linear regression problem (with design matrix ${\bf A} \in \mathbb{R}^{n \times p}$ ) under SPARC
regularization  is formulated as
\vspace{-.2cm}\begin{equation} \label{eq:sparc}
\min_{{\bf x} \in \mathbb{R}^p} \frac{1}{2} \Vert{\bf y} - {\bf A}{\bf x}\Vert^2 + \overbrace{
\iota_{\Sigma_K}({\bf x}) +
\lambda \!\! \sum_{ i,j \in \Omega_K({\bf x}) ,\:i<j} \max \{ \vert x_i\vert,\vert  x_j\vert \}}^{\phi_{\mbox{\tiny SPARC}}^{\tiny{\lambda, K}}({\bf x})}
\end{equation}
where $\iota_{C}$ denotes the indicator of set $C$ ($\iota_C({\bf x}) = 0$, if ${\bf x}\in C$;
$\iota_C({\bf x}) = +\infty$, if ${\bf x}\not\in C$), $\Sigma_K = \{ {\bf x}: \|{\bf x}\|_0 \leq K\}$
is the set of $K$-sparse vectors, and $\Omega_K({\bf x}) = \mbox{supp}(\mathcal{P}_{\Sigma_K}({\bf x}))$ (where
$\mathcal{P}_{\Sigma_K}({\bf x})$ is the projection on $\Sigma_K$, and $\mbox{supp}({\bf v}) = \{i:\, v_i\neq 0\}$) is the
set of indices of the $K$ largest components of ${\bf x}$ (in magnitude).
This regularizer enforces $K$-sparsity and encourages the non-zeros to be equal in magnitude.

Applying proximal splitting algorithms to address \eqref{eq:sparc} requires
the proximity operator
\begin{equation}\label{POofSPARC}
\mbox{prox}_{\phi_{\mbox{\tiny SPARC}}^{\tiny{\lambda , K}}} ( {\bf v} ) =
\arg\min_{{\bf x} } \Bigl( \phi_{\mbox{\tiny SPARC}}^{\tiny{\lambda, K}} ( {\bf x} )  + \frac{1}{2} \left\|{\bf x}-{\bf v}\right\|^2\Bigr).
\end{equation}
The key observation that allows computing
$\mbox{prox}_{\phi_{\mbox{\tiny SPARC}}^{\tiny{\lambda , K}}} ( {\bf v} )$ is
\[
{\bf x} \in \Sigma_K \; \Rightarrow \; \phi_{\mbox{\tiny SPARC}}^{\tiny{\lambda, K}}({\bf x}) =
\phi_{\mbox{\tiny OSCAR}}^{\tiny{0, \lambda}}\left({\bf x}_{\Omega_K({\bf x})}\right),
\]
where ${\bf x}_{S} \in \mathbb{R}^{|S|}$
is the sub-vector of ${\bf x}$ indexed by an index
subset $S\subseteq \{1,...,p\}$. Combining this with  properties of
proximity operators and ideas from \cite{kyrillidis2012combinatorial} allows showing (naturally,
details are omitted here) that
${\bf z} = \mbox{prox}_{\phi_{\mbox{\tiny SPARC}}^{\tiny{\lambda, K}}}\left({\bf x}\right)$
can be computed as follows:
\[
{\bf z}_{\Omega_K({\bf x})} = \mbox{prox}_{\phi_{\mbox{\tiny OSCAR}}^{\tiny{0, \lambda}}} ( {\bf x}_{\Omega_K({\bf x})} ),\hspace{0.5cm}
{\bf z}_{\overline{\Omega}_K({\bf x})} = \mbox{\boldmath$0$}
\]
where $\mbox{\boldmath$0$}$ is a vector of zeros, $\overline{\Omega}_K({\bf x}) = \{1,...,p\}\setminus \Omega_K({\bf x})$,  and
$\mbox{prox}_{\phi_{\mbox{\tiny OSCAR}}^{\tiny{0, \lambda}}}$ can be obtained using
the algorithm proposed in \cite{zhong2012efficient}. Therefore, we can solve (\ref{eq:sparc}) by proximal
splitting algorithms, such as FISTA \cite{beck2009fast}, TwIST \cite{bioucas2007new}, or SpaRSA \cite{wright2009sparse}, which is the 
algorithm adopted in our experiments. SpaRSA (which stands for {\it sparse reconstruction by separable approximation} \cite{wright2009sparse}) is a fast proximal spltting algorithm, based on the step-length selection method of
Barzilai and Borwein \cite{barzilai1988two}.
Its application to SPARC leads to the following algorithm:
\begin{algorithm}{SpaRSA for solving (1)}{
\label{alg:sparsa}}
Set $k=1$, $\eta > 1$, $\alpha_0 = \alpha_{\tiny \mbox{min}} > 0$, 
$\alpha_{\tiny \mbox{max}} > \alpha_{\tiny \mbox{min}}$,  and  ${\bf x}_0$.\\
      ${\bf v}_0 = {\bf x}_0 -{\bf A}^T \left({\bf A} {\bf x}_0 - {\bf y}\right)/\alpha_0$\\
     ${\bf x}_1  = \mbox{prox}_{\phi_{\mbox{\tiny SPARC}}^{\tiny{\lambda , K}}/\alpha_0} \left({\bf v}_0 \right)$\\
\qrepeat\\
      $\hat{\alpha}_k = \frac{\bigl({\bf x}_k - {\bf x}_{k-1}\bigr)^T {\bf A}^T{\bf A}\bigl({\bf x}_k - {\bf x}_{k-1}\bigr)}{\bigl({\bf x}_k - {\bf x}_{k-1}\bigr)^T\bigl({\bf x}_k - {\bf x}_{k-1}\bigr)}, \;\;\;
		 \alpha_k = \max\left\{\alpha_{\tiny \mbox{min}}, \min\left\{\hat{\alpha}_k,\alpha_{\tiny \mbox{max}}\right\}\right\}$\\
     \qrepeat\\
          ${\bf v}_k = {\bf x}_k -{\bf A}^T \left({\bf A} {\bf x}_k - {\bf y}\right)/\alpha_k$\\
          ${\bf x}_{k+1}  = \mbox{prox}_{\phi_{\mbox{\tiny SPARC}}^{\tiny{\lambda , K}}/\alpha_k} \left({\bf v}_k \right)$\\
		      $\alpha_k \leftarrow \eta \alpha_k$
		 \quntil ${\bf x}_{k+1}$ satisfies an acceptance criterion.\\
     $k \leftarrow k + 1$
\quntil some stopping criterion is satisfied.
\end{algorithm}
In this algorithm, the acceptance criterion in Line 10 may be used to enforce the objective function to decrease; see \cite{wright2009sparse} for details.

\section{Experiments}
In this section, we report results of experiments with  synthetic data and with the breast cancer benchmark data, aimed at comparing the SPARC with the LASSO, EN and OSCAR. In order to measure their performances,
we employ the following six metrics defined on an estimate ${\bf e}$ of an original vector ${\bf x}^*$:
\begin{itemize}
	\item Mean absolute error: $\textbf{MAE} = \left\|{\bf A}({\bf x}^* - {\bf e})\right\|_1$;
	\item Mean square error: $\textbf{MSE} = \left\|{\bf A}({\bf x}^* - {\bf e})\right\|_2^2$;
	\item Selection error rate: $\textbf{SER}=\left\|\left|{\bf x}^*\right|-\left|{\bf e}\right|\right\|_1/p$;
	\item Degrees of freedom (\textbf{DoF}): the number of unique non-zero coefficients of {\bf e};
	\item Classification accuracy (\textbf{CLA}): the number of correct classifications of {\bf e};
	\item Number of non-zero features (\textbf{NNZ}).
\end{itemize}
\subsection{Synthetic data}
we consider a regression problem where ${\bf y} = {\bf A}{\bf x}^* + {\bf w}$,
where the true parameters
\begin{equation} \label{linearmodel}
{\bf x}^* = [\underbrace{3, \cdots, 3}_{15}, \underbrace{0, \cdots, 0}_{25}]^T
\end{equation}
and the design matrix ${\bf A}$ is generated as
\[
\begin{split}
{\bf a}_i &=  {\bf z}_1 + {\epsilon}_i^{\bf x}, {\bf z}_1 \sim \mathcal{N}(0,1), i = 1, \cdots, 5; \\
{\bf a}_i &=  {\bf z}_2 + {\epsilon}_i^{\bf x}, {\bf z}_2 \sim \mathcal{N}(0,1), i = 6, \cdots, 10; \\
{\bf a}_i &= {\bf z}_3 + {\epsilon}_i^{\bf x}, {\bf z}_3 \sim \mathcal{N}(0,1), i = 11, \cdots, 15; \\
{\bf a}_i &\sim \mathcal{N}(0,1), i = 16, \cdots, 40
\end{split}
\]
where ${\epsilon}_i^{\bf x}$ are independent identically distributed $\mathcal{N}(0,0.16), i=1,\cdots,15$.
And then ${\bf A} =  [{\bf a}_1, {\bf a}_2, ..., {\bf a}_{40}]^T$ is further normalized, the noise variance of
${\bf w}$ is 0.01. The number of samples for training,
cross validation and testing are 20, 40 and 200, respectively. Notice that it is an ill-posed training problem, since the number of samples is less than the dimension of ${\bf x}$ ($20<40$).
\begin{figure}[h!]
	\centering
		\includegraphics[width=0.7\columnwidth]{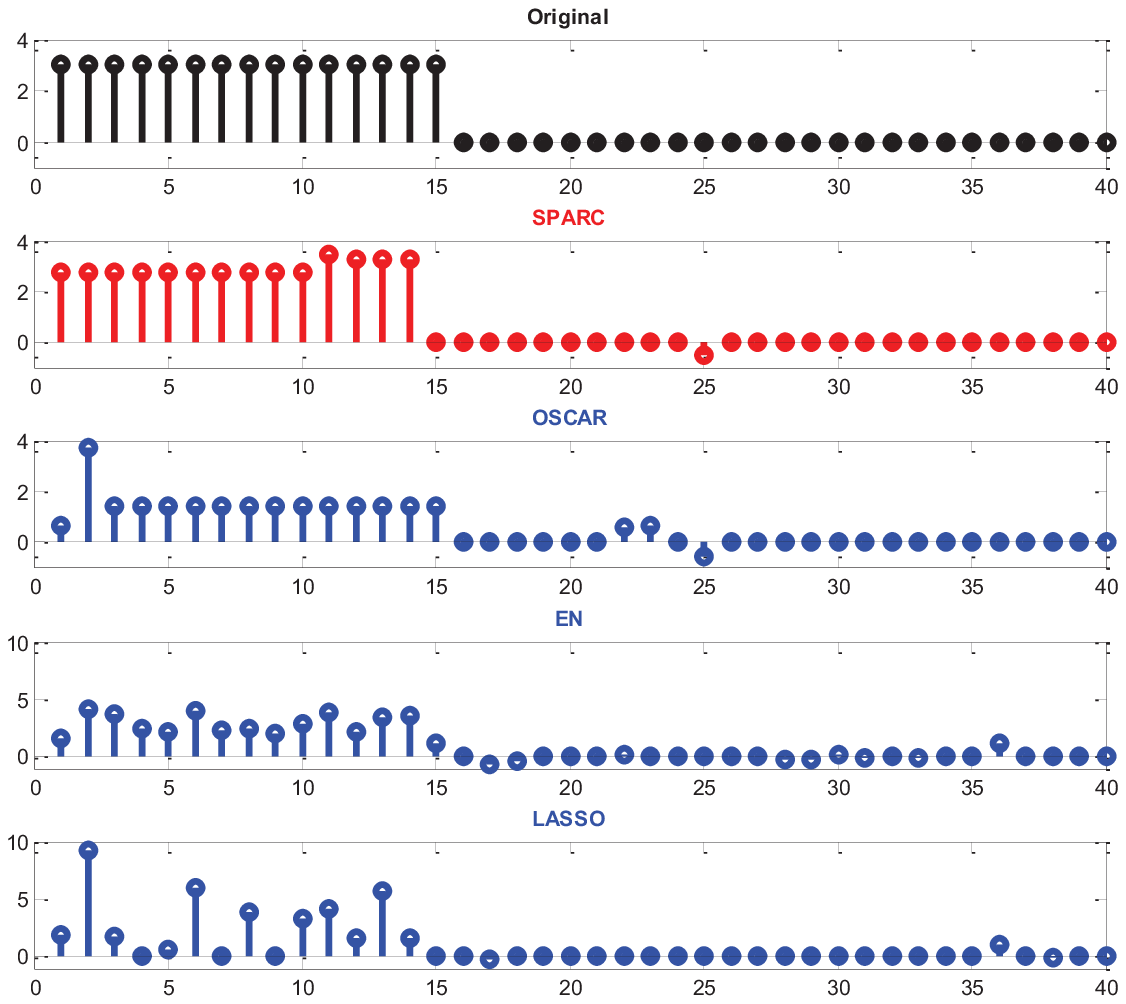}
	\caption{Recovered parameters by different algorithms}
	\label{fig:syntheticdata}
\end{figure}
\begin{table} [h!]
\begin{center}
\begin{tabular}{|l|c|c|c|c|}
\hline
Metrics & LASSO & EN & OSCAR & SPARC \\
\hline\hline
MAE & 27.0677 & 29.4458 & 66.2390 & \textbf{25.7473}\\
MSE & 7.6614 & 7.6939 & 36.8120 & \textbf{5.2904}\\
DoF & 15.64 & 25.28 & 4.56 & \textbf{4.02}\\
SER & 14.50\% & 25.75\% & 8.95\% & \textbf{5.50\%}\\
\hline
\end{tabular}
\end{center}
\caption{Results of the metrics on synthetic data}\label{resultofsyndata}
\end{table}

From Figure \ref{fig:syntheticdata} and Table \ref{resultofsyndata},
the SPARC outperforms the LASSO, EN and OSCAR, showing it is
a promising approach to feature selection and grouping in regression.

\subsection{Breast cancer data}
In this section, we report experiments with the
benchmark breast cancer data, \footnotemark[1]\footnotetext[1]{http://cbio.ensmp.fr/$\scriptsize{\sim}$ljacob/.} which contains 8141 genes in 295 tumors,
where 300 genes that are most correlated with the responses. 50\%, 30\% and 20\% of the data are then randomly chosen for training, cross validation, and testing, respectively. The results averaged over 50 repetitions are show in Table \ref{resultofbredata}.
We can observe that SPARC is a competitive group-sparsity-inducing regularizer for classification
in terms of CLA, and it is able to select features with lower degrees of freedom than LASSO, EN, and OSCAR.
\begin{table}[h!]
\begin{center}
\begin{tabular}{|l|c|c|c|c|}
\hline
Metrics & LASSO & EN & OSCAR & SPARC \\
\hline\hline
CLA & 70.56 & 71.34 & 72.98 & \textbf{74.54}\\
DoF & 41.86 & 180.23 & 39.85 & \textbf{38.12}\\
NNZ & \textbf{41.86} & 180.23 & 120.89 & 80.78\\
\hline
\end{tabular}
\end{center}
\caption{Results of the metrics on breast cancer data} \label{resultofbredata}
\end{table}
\section{Conclusions}
We have proposed the {\it SPARsity and Clustering} (SPARC) regularizer for regression and classification.
We have shown that the proposed SPARC is able to separably enforce $K$-sparsity and encourage the non-zeros to be equal in magnitude, thud accurately grouping the features without parameter shrinkage, outperforming the LASSO, the {\it elastic net}, and the {\it octagonal shrinkage and
clustering algorithm for regression} (OSCAR). Future work will involve considering faster algorithms to solve problems with SPARC regularization.

\bibliography{bibfile}

\begin{thebibliography}{11}
\providecommand{\natexlab}[1]{#1}
\providecommand{\url}[1]{\texttt{#1}}
\expandafter\ifx\csname urlstyle\endcsname\relax
  \providecommand{\doi}[1]{doi: #1}\else
  \providecommand{\doi}{doi: \begingroup \urlstyle{rm}\Url}\fi

\bibitem[Bach et~al.(2012)Bach, Jenatton, Mairal, and
  Obozinski]{bach2012structured}
Francis Bach, Rodolphe Jenatton, Julien Mairal, and Guillaume Obozinski.
\newblock Structured sparsity through convex optimization.
\newblock \emph{Statistical Science}, 27\penalty0 (4):\penalty0 450--468, 2012.

\bibitem[Barzilai and Borwein(1988)]{barzilai1988two}
J.~Barzilai and J.M. Borwein.
\newblock Two-point step size gradient methods.
\newblock \emph{IMA Journal of Numerical Analysis}, 8:\penalty0 141--148, 1988.

\bibitem[Beck and Teboulle(2009)]{beck2009fast}
A.~Beck and M.~Teboulle.
\newblock A fast iterative shrinkage-thresholding algorithm for linear inverse
  problems.
\newblock \emph{SIAM Journal on Imaging Sciences}, 2:\penalty0 183--202, 2009.

\bibitem[Bioucas-Dias and Figueiredo(2007)]{bioucas2007new}
J.M. Bioucas-Dias and M.A.T. Figueiredo.
\newblock A new twist: two-step iterative shrinkage/thresholding algorithms for
  image restoration.
\newblock \emph{IEEE Transactions on Image Processing}, 16:\penalty0
  2992--3004, 2007.

\bibitem[Bondell and Reich(2007)]{bondell2007simultaneous}
H.D. Bondell and B.J. Reich.
\newblock Simultaneous regression shrinkage, variable selection, and supervised
  clustering of predictors with oscar.
\newblock \emph{Biometrics}, 64:\penalty0 115--123, 2007.

\bibitem[Kyrillidis and Cevher(2012)]{kyrillidis2012combinatorial}
Anastasios Kyrillidis and Volkan Cevher.
\newblock Combinatorial selection and least absolute shrinkage via the clash
  algorithm.
\newblock In \emph{Information Theory Proceedings (ISIT), 2012 IEEE
  International Symposium on}, pages 2216--2220. IEEE, 2012.

\bibitem[Tibshirani et~al.(2004)Tibshirani, Saunders, Rosset, Zhu, and
  Knight]{tibshirani2004sparsity}
R.~Tibshirani, M.~Saunders, S.~Rosset, J.~Zhu, and K.~Knight.
\newblock Sparsity and smoothness via the fused lasso.
\newblock \emph{Journal of the Royal Statistical Society (B)}, 67:\penalty0
  91--108, 2004.

\bibitem[Wright et~al.(2009)Wright, Nowak, and Figueiredo]{wright2009sparse}
S.J. Wright, R.D. Nowak, and M.A.T. Figueiredo.
\newblock Sparse reconstruction by separable approximation.
\newblock \emph{IEEE Transactions on Signal Processing}, 57:\penalty0
  2479--2493, 2009.

\bibitem[Yuan and Lin(2005)]{yuan2005model}
M.~Yuan and Y.~Lin.
\newblock Model selection and estimation in regression with grouped variables.
\newblock \emph{Journal of the Royal Statistical Society (B)}, 68:\penalty0
  49--67, 2005.

\bibitem[Zhong and Kwok(2012)]{zhong2012efficient}
L.W. Zhong and J.T. Kwok.
\newblock Efficient sparse modeling with automatic feature grouping.
\newblock \emph{IEEE Transactions on Neural Networks and Learning Systems},
  23:\penalty0 1436--1447, 2012.

\bibitem[Zou and Hastie(2005)]{zou2005regularization}
H.~Zou and T.~Hastie.
\newblock Regularization and variable selection via the elastic net.
\newblock \emph{Journal of the Royal Statistical Society (B)}, 67:\penalty0
  301--320, 2005.

\end{thebibliography}
\end{document}